# Inertial-based Navigation by Polynomial Optimization: Inertial-Magnetic Attitude Estimation

Maoran Zhu, *Student Member, IEEE*, Yuanxin Wu, *Senior Member, IEEE*

*Abstract*— **Inertial-based navigation refers to the navigation methods or systems that have inertial information or sensors as the core part and integrate a spectrum of other kinds of sensors for enhanced performance. Through a series of papers, the authors attempt to explore information blending of inertial-based navigation by a polynomial optimization method. The basic idea is to model rigid motions as finite-order polynomials and then attacks the involved navigation problems by optimally solving their coefficients, taking into considerations the constraints posed by inertial sensors and others. In the current paper, a continuous-time attitude estimation approach is proposed, which transforms the attitude estimation into a constant parameter determination problem by the polynomial optimization. Specifically, the continuous attitude is first approximated by a Chebyshev polynomial, of which the unknown Chebyshev coefficients are determined by minimizing the weighted residuals of initial conditions, dynamics and measurements. We apply the derived estimator to the attitude estimation with the magnetic and inertial sensors. Simulation and field tests show that the estimator has much better stability and faster convergence than the traditional extended Kalman filter does, especially in the challenging large initial state error scenarios.**

## I. INTRODUCTION

Attitude estimation of a rigid body with low-cost MEMS inertial measurement unit (MIMU) sensors has received numerous applications, including but not limited to robotics [1], aerospace [2], unmanned aerial vehicle [3] and human motion tracking [4, 5]. The MIMU usually integrates the three-axis gyroscope, accelerometer and magnetometer into a module or chip. The magnetometer measures the local magnetic field of the Earth, while the gyroscope and accelerometer measure the angular velocity and specific force of the rigid body with respect to the inertial frame, respectively [6].

The inertial and magnetic attitude estimation has witnessed about forty years' endeavors in trying to optimally fuse gyroscope measurements with the valid accelerometer and magnetometer measurements [7, 8]. Generally, the fusion algorithms can be classified into three categories: complementary filter-based methods [9-12], Kalman filter-based methods [7, 13-16] and optimization-based methods [17-19].

The complementary filter [20] utilizes the accelerometer and magnetometer vector measurements to determine the attitude and then combines with the time integration of gyroscope by a proper gain. The complementary filter is a straightforward frequency domain approach with low computation burden. Though quite popular in low cost platforms [10], it does not consider the statistical characteristics of measurements.

As for the Kalman filter-based attitude estimation, the most celebrated tool is the multiplicative extended Kalman filter (EKF) [21], which approximates the attitude nonlinearity by the first-order linearization. However, the local linearization at inaccurate estimate will degrade the filter performance, especially in the case of the large initial state error. The invariant extended Kalman filter [22, 23] utilizes the Lie group affine property and derives the error state model independent of the attitude to improve the stability and robustness of the estimation. To address the linearization issue, the unscented Kalman filter (UKF) is also applied for attitude estimation [24], which employs a set of sigma points to approximate the nonlinear probability distribution and propagates the statistical information of posteriori density. The Monte Carlo-based particle filter is also used for attitude estimation [25] by drawing a large number of random samples to approximate and propagate the probability distribution, which leads to a great computation burden.

There are several early attempts to solve the attitude estimation by the optimization-based method. The works [17, 18, 26] determine the attitude by first optimizing the objective function of the magnetometer and accelerometer measurements and then incorporating gyroscope measurements through a fusion algorithm. Instead of numerically integrating the gyroscope measurements, a more accurate and direct method is to optimize all the available inertial and magnetic measurements simultaneously [19, 27]. The paper [19] proposes the moving-horizon estimation method, which utilizes all measurements in a time window to solve the spacecraft attitude estimation. However, approximating the attitude

This paper was supported in part by National Natural Science Foundation of China (62273228) and Shanghai Jiao Tong University Scientific and Technological Innovation Funds.

Author's address: M. Zhu and Y. Wu are with Shanghai Key Laboratory of Navigation and Location-based Services, School of Electronic Information and Electrical Engineering, Shanghai Jiao Tong University, Shanghai 200240, China (email: zhumaoran@sjtu.edu.cn, yuanx_wu@hotmail.com).

derivative by the first-order difference therein is not accurate enough, especially for the high dynamic applications. A continuous-time batch attitude estimation using B-splines curve to approximate the attitude is proposed in [27], which is not friendly to real time applications.

Inspired by the collocation method, we quite recently come up with a continuous-time state estimation method by the Chebyshev polynomial optimization for general problems [28]. Through a series of papers, we attempt to explore information blending of inertial-based navigation by the polynomial optimization method. In [29], we have introduced the Chebyshev collocation method [30, 31] into the strapdown inertial navigation computation. This paper will further exploit the idea of the collocation method to derive polynomial optimization-based attitude estimators, which transforms the attitude estimation into a problem of Chebyshev coefficient optimization in a sliding window. Specifically, the attitude profile in the current window is represented by the Chebyshev polynomial and the unknown coefficients are then determined by minimizing the weighted residuals of initial conditions, dynamics and measurements. We apply the derived estimators to the attitude and bias estimation with the magnetic and inertial sensors and find that they have much better stability and faster convergence than the traditional EKF does, especially in the large initial state error scenarios.

The rest of this paper is organized as follows. The inertial-magnetic sensor measurement models are first introduced in Section II. Section III derives the polynomial optimization-based inertial-magnetic attitude estimation in terms of the quaternion and Rodrigues vector. A linear initial algorithm is also proposed to provide good initial values for the optimization-based methods. The simulations and field tests are conducted in Section IV. Finally, the conclusion is drawn in Section V.

## II. INERTIAL SENSOR MEASUREMENT MODEL

The gyroscope measures the angular velocity of the body with respect to the inertial axes as

$$\mathbf{y}_g = \boldsymbol{\omega}_{ib}^b + \mathbf{b}_g + \mathbf{n}_g \tag{1}$$

where the subscripts $i$ and $b$ denote the inertial frame ($i$-frame) and body frame ($b$-frame), respectively. $\mathbf{y}_g$ denotes the gyroscope measurement and $\boldsymbol{\omega}_{ib}^b$ denotes the true angular velocity in $b$-frame. $\mathbf{n}_g \sim N\left(0, \mathbf{R}_g\right)$ denotes the gyroscope measurement white noise, and $\mathbf{b}_g$ denotes the gyroscope bias that is modeled as

$$\dot{\mathbf{b}}_g = \mathbf{n}_{bg} \tag{2}$$

where $\mathbf{n}_{bg}$ is Gaussian white noise. When the non-gravitational acceleration such as motion disturbance is small, When the non-gravitational acceleration such as motion disturbance is small, the accelerometer measurement roughly reflects the opposite direction of local gravity in the $b$-frame, i.e.,

$$\mathbf{y}_a \approx -\mathbf{q}_n^{b*} \circ \boldsymbol{\gamma}^n \circ \mathbf{q}_n^b + \mathbf{b}_a + \mathbf{n}_a \tag{3}$$

where the subscript $n$ represents the navigation frame ($n$-frame). Without the loss of generality, the navigation frame in this paper takes the definition of North-Up-East. $\mathbf{n}_a \sim N\left(0, \mathbf{R}_a\right)$ denotes the measurement Gaussian noise and $\mathbf{b}_a$ denotes the measurement noise that is modeled as

$$\dot{\mathbf{b}}_a = \mathbf{n}_{ba} \tag{4}$$

where $\mathbf{n}_{ba}$ is a Gaussian white noise. $\boldsymbol{\gamma}^n = \begin{bmatrix} 0 & -g & 0 \end{bmatrix}^T$ denotes the gravity vector in $n$-frame and $g$ denotes the magnitude of the local gravity. $\mathbf{q}_n^b \triangleq \begin{bmatrix} s & \boldsymbol{\eta}^T \end{bmatrix}^T$ encodes the unit attitude quaternion of $b$-frame relative to $n$-frame, where $s$ is the scale part and $\boldsymbol{\eta}$ is the vector part of $\mathbf{q}_n^b$. The operator $\circ$ denotes the quaternion multiplication, defined as

$$\mathbf{q}_1 \circ \mathbf{q}_2 = \begin{bmatrix} \overset{+}{\mathbf{q}}_1 \end{bmatrix} \begin{bmatrix} s_2 \\ \boldsymbol{\eta}_2 \end{bmatrix} = \begin{bmatrix} \overset{-}{\mathbf{q}}_2 \end{bmatrix} \begin{bmatrix} s_1 \\ \boldsymbol{\eta}_1 \end{bmatrix} \tag{5}$$

The two quaternion multiplication matrices, $\begin{bmatrix} \overset{+}{\mathbf{q}} \end{bmatrix}$ and $\begin{bmatrix} \overset{-}{\mathbf{q}} \end{bmatrix}$, are respectively defined by

$$\begin{bmatrix} \overset{+}{\mathbf{q}} \end{bmatrix} \triangleq \begin{bmatrix} s & -\boldsymbol{\eta}^T \\ \boldsymbol{\eta} & s\mathbf{I}_3 + \boldsymbol{\eta} \times \end{bmatrix}, \quad \begin{bmatrix} \overset{-}{\mathbf{q}} \end{bmatrix} \triangleq \begin{bmatrix} s & -\boldsymbol{\eta}^T \\ \boldsymbol{\eta} & s\mathbf{I}_3 - \boldsymbol{\eta} \times \end{bmatrix} \tag{6}$$

where $\mathbf{I}_n$ denotes an $n \times n$ identity matrix. The skew symmetric matrix $(\cdot \times)$ is defined that the cross product $x \times y = (x \times) y$ is satisfied for arbitrary two vectors. The conjugate of the unit quaternion $\mathbf{q}_n^{b*}$ in (3) is defined as $\mathbf{q}_n^{b*} = \begin{bmatrix} s & -\boldsymbol{\eta}^T \end{bmatrix}^T$.

The magnetometer triad measures the total magnetic flux density in $b$-frame, given as

$$\mathbf{y}_m = \mathbf{q}_n^{b*} \circ \mathbf{m}^n \circ \mathbf{q}_n^b + \mathbf{n}_m \tag{7}$$

where $\mathbf{y}_m$ and $\mathbf{n}_m \sim N(0, \mathbf{R}_m)$ denote the normalized magnetometer triad measurements (unit norm) and its noise, respectively. $\mathbf{m}^n$ is the normalized magnetic field in $n$-frame, expressed as

$$\mathbf{m}^n = \begin{bmatrix} \cos\alpha_m \cos\gamma_m \\ -\sin\gamma_m \\ \sin\alpha_m \cos\gamma_m \end{bmatrix} \tag{8}$$

where $\alpha_m$ and $\gamma_m$ are, respectively, the magnetic declination and inclination [6]. It is noted that other sensor parameters, such as the scale factor and misalignment, are assumed having been well calibrated [32, 33]. In this paper, the accelerometer and magnetometer measurements are only used when the external acceleration and magnetic disturbance retain small. Specifically, the norm-based detectors are applied to select the feasible measurements [16]

$$\left| \|\mathbf{y}_a\| - g \right| < \varepsilon_a \tag{9}$$

$$\left| 1 - \|\mathbf{y}_m\| \right| < \varepsilon_m \tag{10}$$

where the first inequality represents the accelerometer detector and the second inequality represents the magnetometer detector,

$\|\cdot\|$ denotes the norm of a vector, and $\varepsilon_a$ and $\varepsilon_m$ denote the given thresholds of the two detectors.

Considering the Earth rotation rate, the dynamics of the attitude quaternion satisfies [6]

$$\dot{\mathbf{q}}_n^b = \mathbf{q}_n^b \circ \boldsymbol{\omega}_{nb}^b / 2 = \left( \mathbf{q}_n^b \circ \boldsymbol{\omega}_{ib}^b - \boldsymbol{\omega}_{in}^n \circ \mathbf{q}_n^b \right) / 2 \qquad (11)$$

where $\boldsymbol{\omega}_{nb}^b$ denotes the body angular velocity vector with respect to $n$-frame expressed in $b$-frame. $\boldsymbol{\omega}_{in}^n \approx \boldsymbol{\omega}_{ie}^n$ for the low-speed vehicles and $\boldsymbol{\omega}_{ie}^n = \begin{bmatrix} \Omega \cos L & \Omega \sin L & 0 \end{bmatrix}^T$ denotes the Earth's angular velocity vector expressed in $n$-frame, among which $\Omega$ is the norm of the Earth rotation and $L$ is the latitude of the location. Substituting (1) into (11) and multiplying the conjugate of $\mathbf{q}_n^b$ on both sides, the angular velocity measurements can be represented as a function of quaternion and gyroscope bias, as

$$\mathbf{y}_g = 2\mathbf{q}_n^{b*} \circ \dot{\mathbf{q}}_n^b + \mathbf{q}_n^{b*} \circ \boldsymbol{\omega}_{ie}^n \circ \mathbf{q}_n^b + \mathbf{b}_g + \mathbf{n}_g \qquad (12)$$

Another popular parameter to present the attitude is the Rodrigues vector. It has only three components and no inherent constraint, but is singular for a $180°$ rotation [21]. Therefore, the Rodrigues vector is usually used to parameterize the attitude update in a short time interval. In this regard, the attitude in the time interval can be written as

$$\mathbf{q}_n^b(\tau) = \mathbf{q}_{n,0}^b \circ \Delta \mathbf{q}(\tau) \qquad (13)$$

where $\mathbf{q}_{n,0}^b$ denotes the initial attitude quaternion at the start of the current time window and $\Delta \mathbf{q}$ denotes the quaternion update, which is related with the Rodrigues vector update as

$$\Delta \mathbf{q} = \frac{\begin{bmatrix} 2 & \Delta \mathbf{g}^T \end{bmatrix}^T}{\sqrt{4 + \|\Delta \mathbf{g}\|^2}} \qquad (14)$$

reversely,

$$\Delta \mathbf{g} = \frac{2 \Delta \boldsymbol{\eta}}{\Delta s} \qquad (15)$$

where $\Delta s$ is the scale part and $\Delta \boldsymbol{\eta}$ is the vector part of $\Delta \mathbf{q}$. Substituting (13) and (14) into (12), the angular velocity measurements can also be represented by the Rodrigues vector update

$$\mathbf{y}_g = \frac{4 \Delta \dot{\mathbf{g}} - 2 \Delta \mathbf{g} \times \Delta \dot{\mathbf{g}}}{4 + \|\Delta \mathbf{g}\|^2} + \Delta \mathbf{C}(\Delta \mathbf{g}) \mathbf{C}_{n,0}^b \boldsymbol{\omega}_{ie}^n + \mathbf{b}_g + \mathbf{n}_g \qquad (16)$$

where $\mathbf{C}_{n,0}^b$ is the initial rotation matrix corresponding to the initial attitude quaternion $\mathbf{q}_{n,0}^b$ and $\Delta \mathbf{C}(\Delta \mathbf{g})$ means a function that transforms the Rodrigues vector to the corresponding rotation matrix.

## III. INERTIAL-MAGNETIC ATTITUDE ESTIMATION BY CHEBYSHEV POLYNOMIAL OPTIMIZATION (AttEstPO)

### A. Formulation of Attitude Estimation

Without the loss of the generality, consider the attitude estimation on the time interval $\begin{bmatrix} t_0 & t_M \end{bmatrix}$, in which $M+1$ inertial-magnetic samples are available at $t_k$, $k = 0, \cdots M$. Because the gyroscope and accelerometer biases are slowly changing, it is reasonable to treat them as constants in a short time window. Therefore, the state to be estimated in the time interval includes the continuous attitude and the constant gyroscope and accelerometer biases, which are denoted as $\mathbf{x}(t) \triangleq \begin{bmatrix} \mathbf{q}_n^b(t)^T, & \mathbf{b}_a^T, & \mathbf{b}_g^T \end{bmatrix}^T$. Assume the initial state is given as $\mathbf{x}_0 = \begin{bmatrix} \mathbf{q}_{n,0}^{b \, T}, & \mathbf{b}_{a_0}^T, & \mathbf{b}_{g_0}^T \end{bmatrix}^T$, where $\mathbf{q}_{n,0}^b$, $\mathbf{b}_{a_0}$ and $\mathbf{b}_{g_0}$ denote the quaternion, accerlerometer and gyroscope biases at the initial time $t_0$, respectively.

In the least squares sense [34], an optimal continuous-discrete attitude estimation is to minimize the measurement residuals with respect to $\mathbf{x}(t)$

$$\min_{\mathbf{x}(t)} \left( J_{\mathbf{x}_0} + J_{\mathbf{v}} + J_{\mathbf{z}} \right) \quad \text{s.t } \|\mathbf{q}(t)\| = 1 \qquad (17)$$

where the objective functions, $J_{\mathbf{x}_0}$, $J_{\mathbf{v}}$ and $J_{\mathbf{z}}$, respectively denote the prior, dynamics and measurement terms. They are explicitly given as follows

$$
\begin{aligned}
J_{\mathbf{x}_0} &= \mathbf{e}_{\mathbf{x}_0}^T \mathbf{e}_{\mathbf{x}_0} \\
J_{\mathbf{v}} &= \int_{t_0}^{t_M} \mathbf{e}_{\mathbf{v}}^T(\tau) \mathbf{e}_{\mathbf{v}}(\tau) d\tau \\
J_{\mathbf{z}} &= \sum_{k=1}^p \mathbf{e}_{a_k}^T \mathbf{e}_{a_k} + \sum_{k=1}^s \mathbf{e}_{m_k}^T \mathbf{e}_{m_k}
\end{aligned}
\qquad (18)
$$

where $p$ and $s$ respectively denote the number of valid accelerometer and magnetometer measurements in the current time window that are selected by the detectors in (9) and (10). With the measurement models (3), (7) and (12), the weighted residuals $\mathbf{e}_{\mathbf{x}_0}$, $\mathbf{e}_{\mathbf{v}}$, $\mathbf{e}_a$ and $\mathbf{e}_m$ are given as

$$
\begin{aligned}
\mathbf{e}_{\mathbf{x}_0} &= \mathbf{W}_{\mathbf{x}_0}^T \begin{bmatrix} \delta \boldsymbol{\psi}_0^T & \mathbf{b}_a^T - \mathbf{b}_{a_0}^T & \mathbf{b}_g^T - \mathbf{b}_{g_0}^T \end{bmatrix} \\
\mathbf{e}_{\mathbf{v}} &= \mathbf{W}_{\mathbf{v}}^T \left( \mathbf{y}_g - 2\mathbf{q}_n^{b*} \circ \dot{\mathbf{q}}_n^b - \mathbf{q}_n^{b*} \circ \boldsymbol{\omega}_{ie}^n \circ \mathbf{q}_n^b - \mathbf{b}_g \right) \\
\mathbf{e}_a &= \mathbf{W}_a^T \left( \mathbf{y}_a + \mathbf{q}_n^{b*} \circ \mathbf{g}^n \circ \mathbf{q}_n^b - \mathbf{b}_a \right) \\
\mathbf{e}_m &= \mathbf{W}_m^T \left( \mathbf{y}_m - \mathbf{q}_n^{b*} \circ \mathbf{m}^n \circ \mathbf{q}_n^b \right)
\end{aligned}
\qquad (19)
$$

which are related to the prior, attitude dynamics and valid accelerometer and magnetometer measurements, respectively, through their corresponding weight matrices. Specifically, regarding the residual of initial state, $\delta \boldsymbol{\psi}_0$ is the initial three-dimensional attitude error that is roughly related to the attitude quaternion as [6]

$$\delta \boldsymbol{\psi} \approx 2 \begin{bmatrix} \mathbf{q}_n^b \circ \hat{\mathbf{q}}_n^{b*} \end{bmatrix}_{2:4} \qquad (20)$$

where $\mathbf{q}_n^b$ and $\hat{\mathbf{q}}_n^b$ denote the true and error-contaminated quaternions, respectively. The operator $\begin{bmatrix} \bullet \end{bmatrix}_{2:4}$ extracts the second to the fourth rows of a matrix. The weight matrices are obtained from the Cholesky factorization of the inverse covariance matrix, namely, $\mathbf{P}_{\mathbf{x}_0}^{-1} = \mathbf{W}_{\mathbf{x}_0} \mathbf{W}_{\mathbf{x}_0}^T$, $\mathbf{R}_v^{-1} = \mathbf{W}_g \mathbf{W}_g^T$, $\mathbf{R}_a^{-1} = \mathbf{W}_a \mathbf{W}_a^T$, $\mathbf{R}_m^{-1} = \mathbf{W}_m \mathbf{W}_m^T$. It should be noted that the covariance matrix of the initial state is defined as $\mathbf{P}_{\mathbf{x}_0}^{-1} \triangleq diag\left[\mathbf{P}_{\mathbf{\psi}_0}, \mathbf{P}_{b_{a},0}, \mathbf{P}_{b_{g},0}\right]^{-1}$, where $\mathbf{P}_{\mathbf{\psi}_0}$, $\mathbf{P}_{b_{a},0}$ and $\mathbf{P}_{b_{g},0}$ are the initial covariances of the attitude, accelerometer and gyroscope biases, respectively.

### B. Attitude Estimation by Chebyshev Polynomial Optimization

To solve the infinite-dimension optimization problem in (17), the Chebyshev collocation method is introduced to represent the attitude by a finite-order Chebyshev polynomial and transform the continuous-time attitude estimation into a constant parameter optimization problem.

The Chebyshev polynomial of the first kind is defined over the internal $[-1 \quad 1]$ by the recurrence relation as [30]

$$F_0(\tau) = 1, \ F_1(\tau) = \tau,$$
$$F_{i+1}(\tau) = 2\tau F_i(\tau) - F_{i-1}(\tau) \ \text{ for } i \geq 1 \quad (21)$$

where $F_i(\tau)$ is the $i^{\text{th}}$-degree Chebyshev polynomial. In order to apply the Chebyshev polynomial, the optimization in (17) on $t \in [t_0 \quad t_M]$ needs to be mapped into $\tau \in [-1 \quad 1]$ by the affine transformation as follows

$$\tau = \frac{2}{t_M - t_0} t - \frac{t_M + t_0}{t_M - t_0} \quad (22)$$

The attitude quaternion in the time interval is then approximated by a Chebyshev polynomial up to order $N_q$ as

$$\mathbf{q}_n^b(\tau) \approx \sum_{i=0}^{N_q} \mathbf{d}_i F_i(\tau) \triangleq \mathbf{DF}(\tau) \quad (23)$$

where $\mathbf{d}_i$ denote the $i^{\text{th}}$-degree Chebyshev coefficient. The matrices $\mathbf{F}(\tau) \triangleq \left[F_0(\tau) \ F_1(\tau) \ \cdots \ F_{N_q}(\tau)\right]^T$ and $\mathbf{D} \triangleq \left[\mathbf{d}_0, \cdots, \mathbf{d}_{N_q}\right]$ are defined for compact denotation.

Then, the derivative of the attitude quaternion is given by

$$\dot{\mathbf{q}}_n^b(\tau) \approx \sum_{i=0}^{N_q} \mathbf{d}_i \dot{F}_i(\tau) \triangleq \mathbf{D}\dot{\mathbf{F}}(\tau) \quad (24)$$

where $\dot{\mathbf{F}}(\tau) \triangleq \left[\dot{F}_0(\tau) \ \dot{F}_1(\tau) \ \cdots \ \dot{F}_{N_q}(\tau)\right]^T$ and $\dot{F}_i$ is the time derivative of $F_i$ that is directly obtained by taking the derivative of (21). That is to say

$$\dot{F}_0(\tau) = 0, \ \dot{F}_1(\tau) = 1,$$
$$\dot{F}_{i+1}(\tau) = 2F_i(\tau) + 2\tau \dot{F}_i(\tau) - \dot{F}_{i-1}(\tau) \ \text{ for } i \geq 1 \quad (25)$$

The reason that the Chebyshev polynomial is selected as the basis function is for its high accuracy and efficacy in functional approximation. As a matter of fact, the Chebyshev polynomial is very close to the best polynomial approximation in the $\infty$-norm [30]. The integral term $J_v$ in (18) can be numerically solved by the Clenshaw-Curtis quadrature formula as [30]

$$J_v \approx \sum_{i=0}^{N} w_i \mathbf{e}_v^T(\tau_i) \mathbf{e}_v(\tau_i) \quad (26)$$

where $\tau_i$ denotes the Chebyshev points

$$\tau_i = -\cos(i\pi/N), \quad i = 0, 1, \cdots, N \quad (27)$$

and $w_i$ denotes the weight that is determined by the integrals of the Lagrange polynomials [35] and $N+1$ is the number of Chebyshev points. We see that (26) requires the angular velocity at the Chebyshev points. In this paper, it is achieved by the way of the extended Floater and Hormann (EFH) interpolation method [36], which reconstructs the angular velocity from the equally-spaced time sampled angular velocity measurements [37]. Readers are referred to [29] for the details of the angular velocity reconstruction.

Following the commonly used strategy to handle the constraints in the collocation-based optimal control [38], the continuous unit quaternion constraint in (17) is discretized at the Chebyshev points as

$$\left\|\mathbf{q}_n^b(\tau_k)\right\| = 1 \quad k = 0, 1, \cdots, N \quad (28)$$

Substituting the Chebyshev approximations (23), (24), (26) and the discretized constraints (28) into (17), we can reformulate the estimation task for the time window of interest as

$$\min_{\{\mathbf{D}, \mathbf{b}_a, \mathbf{b}_g\}} \left(J_{\mathbf{x}_0} + J_v + J_z\right)$$
$$\text{s.t. } \left\|\mathbf{q}_n^b(\tau_k)\right\| = 1 \quad k = 0, 1, \cdots, N \quad (29)$$

where the estimation parameters include the Chebyshev coefficient and the accelerometer/gyroscope biases. Equation (29) is a constrained nonlinear squares problem, which can be transformed to an unconstrained nonlinear least squares by the augmented Lagrangian method and then solved by the Levenberg-Marquardt algorithm [39]. Once the coefficients are determined, the continuous attitude quaternion as a function of time will be finally acquired by (23).

It should be noted that (29) is a constrained optimization that is generally more time-consuming than an unconstrained optimization. On the other hand, the above optimization problem can be reformulated as an unconstrained one using the Rodrigues vector. To avoid its singularity at a $180°$ rotation, however, we consider the Rodrigues vector update in the short-time window. Specifically, the Rodrigues vector update is approximated by a Chebyshev polynomial up to order $N_g$ as

$$\Delta \mathbf{g}(\tau) \approx \sum_{i=0}^{N_g} \mathbf{h}_i F_i(\tau) \triangleq \mathbf{HF}(\tau) \quad (30)$$

where $\mathbf{h}_i$ denote the $i^{\text{th}}$-degree Chebyshev coefficient, and $\mathbf{H} \triangleq \left[\mathbf{h}_0, \cdots, \mathbf{h}_{N_g}\right]$ denotes the combined matrix of Chebyshev coefficients. Then, the derivative of the attitude quaternion is given by

$$\Delta \dot{\mathbf{g}}(\tau) \approx \sum_{i=0}^{N_g} \mathbf{h}_i \dot{F}_i(\tau) \triangleq \mathbf{H}\dot{\mathbf{F}}(\tau) \quad (31)$$

By analogy with the quaternion estimation, the Rodrigues vector update estimation can be formulated as

$$\min_{\{\mathbf{H}, \mathbf{b}_a, \mathbf{b}_g\}} \left(J_{\mathbf{x}_0} + J_v + J_z\right) \quad (32)$$

where the estimation parameters include the Chebyshev coefficient and the accelerometer/magnetometer biases. The weighted residuals of initial state and accelerometer/magnetometer measurements are obtained by substituting (13) and (14) into (19), and the weighted residual of attitude dynamics is obtained from (16), given as

$$\mathbf{e}_v = \mathbf{W}_g^T \left( \mathbf{y}_g - \frac{4\Delta\dot{\mathbf{g}} - 2\Delta\mathbf{g}\times\Delta\dot{\mathbf{g}}}{4+\|\Delta\mathbf{g}\|^2} - \Delta\mathbf{C}(\Delta\mathbf{g})\mathbf{C}_{n,0}^b\boldsymbol{\omega}_{ie}^n - \mathbf{b}_g \right) \quad (33)$$

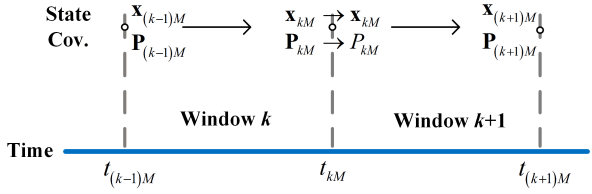

Fig. 1. Estimated state and its covariance propagation (the state and covariance at the end of Window $k$ are served as the initial condition for Window $k+1$).

With the optimized Chebyshev coefficients, the attitude quaternion in the window is readily obtained by substituting (30) into (13) and (14).

In summary, the approximations involved in the proposed batch estimators include the attitude profile approximation by the Chebyshev polynomial in (23) or (30), the Clenshaw-Curtis quadrature in (26) and the angular velocity interpolant by the EFH strategy. With the increased order of Chebyshev polynomial, the errors incurred by Chebyshev approximation and Clenshaw-Curtis quadrature will decay towards zero [30]. Additionally, the EFH interpolant is also likely very close to optimality for time-equispaced samples [40]. In this regard, the proposed attitude estimator by polynomial optimization (AttEstPO) is nearly optimal in the least squares sense.

To meet the requirement of real-time applications, the above optimizations are expected to be solved recursively on consecutive short-time windows. The estimation result at the end of the current window can serve as the prior for the next window, but the required covariance estimation is not directly available. In view of the strategy in the moving horizon estimation [41], the multiplicative EKF, with the linearized dynamics and measurements at the AttEstPO estimate, can be resorted to compute the state covariance. The multiplicative EKF, also known as the error-state EKF, estimates the error state (instead of the original state) to avoid the over-parameterization issue of the attitude [6]. The error state is defined as the estimate subtracting the truth, i.e., $\delta\mathbf{x} = \hat{\mathbf{x}} - \mathbf{x}$, except that the definition of attitude error follows (20). The linearized error-state dynamics is expressed as

$$\delta\dot{\mathbf{x}} = \mathbf{B}\delta\mathbf{x} + \mathbf{G}\mathbf{w} \quad (34)$$

where the error state is a 9-dimensional vector $\delta\mathbf{x} \triangleq \begin{bmatrix} \delta\boldsymbol{\psi} & \delta\mathbf{b}_a & \delta\mathbf{b}_g \end{bmatrix}^T$, the dynamic noise is $\mathbf{w} \triangleq \begin{bmatrix} \mathbf{n}_g & \mathbf{n}_{ba} & \mathbf{n}_{bg} \end{bmatrix}^T$ and the matrices are

$$\mathbf{B} = \begin{bmatrix} \mathbf{0}_3 & \mathbf{0}_3 & -\mathbf{C}_b^n \\ \mathbf{0}_3 & \mathbf{0}_3 & \mathbf{0}_3 \\ \mathbf{0}_3 & \mathbf{0}_3 & \mathbf{0}_3 \end{bmatrix} \quad \mathbf{G} = \begin{bmatrix} -\mathbf{C}_b^n & \mathbf{0}_3 & \mathbf{0}_3 \\ \mathbf{0}_3 & \mathbf{I}_3 & \mathbf{0}_3 \\ \mathbf{0}_3 & \mathbf{0}_3 & \mathbf{I}_3 \end{bmatrix} \quad (35)$$

where $\mathbf{0}_n$ denotes an $n\times n$ zero matrix. The multiplicative EKF covariance prediction from $t_{k-1}$ to $t_k$ is given as

$$\mathbf{P}_k^- = (\mathbf{I}_9 + \mathbf{B}_k)\mathbf{P}_{k-1}(\mathbf{I}_9 + \mathbf{B}_k)^T + \mathbf{G}_k\mathbf{Q}_k\mathbf{G}_k^T T \quad (36)$$

where the covariance update interval is set to $T \triangleq t_k - t_{k-1}$. When the measurement at time $t_k$ comes, the covariance is updated by

$$\mathbf{P}_k = \mathbf{P}_k^- - \mathbf{P}_k^- \mathbf{H}_k^T \left( \mathbf{H}_k \mathbf{P}_k^- \mathbf{H}_k^T + \mathbf{R}_k \right)^{-1} \mathbf{H}_k \mathbf{P}_k^- \quad (37)$$

where $\mathbf{R}_k$ and $\mathbf{H}_k$ denote the measurement noise covariance and linearized measurement matrices, respectively. For the accelerometer and magnetometer, the matrix $\mathbf{H}_k$ is respectively written as

$$\begin{aligned} \mathbf{H}_{k,m} &= \begin{bmatrix} \mathbf{C}_n^b\mathbf{m}^n\times & \mathbf{0}_3 & \mathbf{0}_3 \end{bmatrix} \\ \mathbf{H}_{k,a} &= \begin{bmatrix} -\mathbf{C}_n^b\boldsymbol{\tau}^n\times & \mathbf{I}_3 & \mathbf{0}_3 \end{bmatrix} \end{aligned} \quad (38)$$

Note that the matrices $\mathbf{G}_k$, $\mathbf{H}_k$ and $\mathbf{B}_k$ in (36) and (37) are calculated using the AttEstPO estimate. In contrast, the covariance propagation of EKF is performed at the current state estimate, which is usually inferior to that by AttEstPO because the latter is a kind of local smoothing using all information during the current time window and EKF only uses information up to the current time. Taken as an example, Fig. 1 illustrates the state and covariance propagation between two adjacent windows, where the state estimate and its covariance at the end time of window $k$ serve as the initial condition for window $k+1$.

From the theoretical perspective, the proposed AttEstPO algorithm only makes a Gaussian assumption of the prior and does not introduce any approximations of nonlinear dynamics and measurements in the time interval of interest. This feature is a significant advantage over the well-known linearization-based EKF that approximates the nonlinearity by successive Taylor expansion at current estimate. In this regard, the AttEstPO is promising to obtain a better estimation at the price of time delay, as shown in Section IV.

### C. Chebyshev Coefficient Initialization

A fine initialization of Chebyshev coefficients is required for the above nonlinear optimization. In this subsection, we present a linear algorithm to acquire the initial attitude Chebyshev coefficients by assuming approximately known biases and correct initial attitude. Multiplying $\mathbf{q}_n^b$ on both sides of the attitude dynamics (12) and the accelerometer/magnetometer measurement in (3) and (7), ignoring the measurement noises and using the quaternion multiplication property in (5), we have

$$\begin{aligned} \boldsymbol{\rho}_a(t)\mathbf{q}_n^b(t) &= \mathbf{0}_{4\times1} \\ \boldsymbol{\rho}_m(t)\mathbf{q}_n^b(t) &= \mathbf{0}_{4\times1} \\ \boldsymbol{\rho}_g(t)\mathbf{q}_n^b(t) - 2\dot{\mathbf{q}}_n^b(t) &= \mathbf{0}_{4\times1} \end{aligned} \quad (39)$$

where the matrices are readily defined as

$$\begin{aligned} \boldsymbol{\rho}_a(t) &\triangleq \left[ \left( \mathbf{y}_a(t) - \mathbf{b}_a \right) \right]^+ + \left[ \boldsymbol{\gamma}^n \right]^+ \\ \boldsymbol{\rho}_m(t) &\triangleq \left[ \mathbf{y}_m(t) \right]^- - \left[ \mathbf{m}^n \right]^+ \\ \boldsymbol{\rho}_g(t) &\triangleq \left[ \left( \mathbf{y}_g(t) - \mathbf{b}_g \right) \right]^- - \left[ \boldsymbol{\omega}_{ie}^n(t) \right]^+ \end{aligned} \quad (40)$$

TABLE I
INERTIAL-MAGNETIC ATTITUDE ESTIMATION BY POLYNOMIAL OPTIMIZATION

| | **Qua-AttEstPO** | **Rod-AttEstPO** |
|---|---|---|
| Input: | Chebyshev order $N_q$ or $N_g$, initial state $\mathbf{x}_0$ with associated covariance $\mathbf{P}_0$, inertial and magnetic measurements $\mathbf{y}_g(t_i), \mathbf{y}_a(t_i), \mathbf{y}_m(t_i)$ $i=0,1,\ldots,M$ and the measurement noise covariances $\mathbf{R}_g, \mathbf{R}_a, \mathbf{R}_m$ | |
| Step 1: | Interpolate angular velocity measurement by EFH and obtain angular velocity at Chebyshev points $\mathbf{y}_g(\tau_i)$ $i=0,1,\cdots,N$ | |
| Step 2: | Compute Chebyshev polynomial coefficients of quaternion $\mathbf{q}_m^b$ by homogenous least squares (Eq. (45)) $\min_{\mathbf{D}}\left\|\mathbf{A}\,vec(\mathbf{D})\right\|$ s.t $\left\|\mathbf{B}\,vec(\mathbf{D})\right\|=1$ | |
| Step 3: | Obtain quaternion Chebyshev coefficients and sensor biases by solving constrained nonlinear least squares (Eq. (29)) $\min_{\{\mathbf{D},\mathbf{b}_a,\mathbf{b}_g\}}\left(J_{\mathbf{x}_0}+J_v+J_z\right)$ s.t. $\left\|\mathbf{q}_n^b(\tau_k)\right\|=1$ $k=0,1,\cdots,N$ | Acquire initial coefficients of the Rodrigues vector $\Delta\mathbf{g}$ using quaternion Chebyshev coefficients (Eq. (47)) $\mathbf{h}_i \approx \dfrac{2-\delta_{0i}}{P_{N_g}}\sum_{k=0}^{P_{N_g}-1}\Delta\mathbf{g}\left(\Delta\mathbf{q}\left(\cos\dfrac{(k+1/2)\pi}{P_{N_g}}\right)\right)\cos\dfrac{i(k+1/2)\pi}{P_{N_g}}$ Compute Rodrigues vector Chebyshev polynomial coefficients and sensor biases by nonlinear least squares (Eq. (32)) $\min_{\{\mathbf{H},\mathbf{b}_a,\mathbf{b}_g\}}\left(J_{\mathbf{x}_0}+J_v+J_z\right)$ |
| Step 4: | Obtain attitude quaternion (Eq. (23)) $\mathbf{q}_n^b(\tau)=\sum_{i=0}^{N_q}\mathbf{d}_i F_i(\tau)$ | Obtain Rodrigues vector update and transform it to quaternion (Eqs. (30), (13) and (14)) $\Delta\mathbf{g}(\tau)=\sum_{i=0}^{N_g}\mathbf{h}_i F_i(\tau)$ $\mathbf{q}_n^b(\tau)=\mathbf{q}_{n,0}^b\circ\Delta\mathbf{q}\left(\Delta\mathbf{g}(\tau)\right)$ |
| Step 5: | Compute estimate covariance by multiplicative EKF using AttEstPO estimated attitude (Eqs. (36)-(37)) $\mathbf{P}_k^-=\left(\mathbf{I}_9+\mathbf{B}_k\right)\mathbf{P}_{k-1}\left(\mathbf{I}_9+\mathbf{B}_k\right)^T+\mathbf{G}_k\mathbf{Q}_k\mathbf{G}_k^T T$ $\mathbf{P}_k=\mathbf{P}_k^- - \mathbf{P}_k^-\mathbf{H}_k^T\left(\mathbf{H}_k\mathbf{P}_k^-\mathbf{H}_k^T+\mathbf{R}_k\right)^{-1}\mathbf{H}_k\mathbf{P}_k^-$ | |

Substituting the quaternion Chebyshev polynomial (23) into (39) yields

$$\mathbf{a}_a(t)vec(\mathbf{D})=\mathbf{0}_{4\times1}$$
$$\mathbf{a}_m(t)vec(\mathbf{D})=\mathbf{0}_{4\times1} \quad (41)$$
$$\mathbf{a}_g(t)vec(\mathbf{D})=\mathbf{0}_{4\times1}$$

where $vec(\mathbf{D})$ is formed by stacking the columns of $\mathbf{D}$ and the involved matrices are defined as

$$\mathbf{a}_a(t)\triangleq\left[\mathbf{F}^T(t)\otimes\boldsymbol{\rho}_a(t)\right]$$
$$\mathbf{a}_m(t)\triangleq\left[\mathbf{F}^T(t)\otimes\boldsymbol{\rho}_m(t)\right] \quad (42)$$
$$\mathbf{a}_g(t)\triangleq\left[\mathbf{F}^T(t)\otimes\boldsymbol{\rho}_g(t)-2\dot{\mathbf{F}}^T(t)\otimes\mathbf{I}_4\right]$$

where $\otimes$ denotes the Kronecker product. Note that the identity property $vec(\mathbf{A}_1\mathbf{A}_2\mathbf{A}_3)=\left(\mathbf{A}_3^T\otimes\mathbf{A}_1\right)vec(\mathbf{A}_2)$ has been used in deriving (41), where $\mathbf{A}_1$, $\mathbf{A}_2$, $\mathbf{A}_3$ are matrices of appropriate dimensions. Using the quaternion multiplication property and the definition of the attitude error in (20), a correct initial attitude means

$$\delta\boldsymbol{\psi}_0\approx2\left[\mathbf{q}_n^b(\tau_0)\otimes\mathbf{q}_{n,0}^{b}{}^{*}\right]_{2:4}=2\left[\left[\mathbf{q}_{n,0}^{b}{}^{-}\right]\mathbf{q}_n^b(\tau_0)\right]_{2:4}=\mathbf{0}_{3\times1} \quad (43)$$

Approximating $\mathbf{q}_n^b$ by Chebyshev polynomial, (43) is written as

$$\mathbf{A}_{init}vec(\mathbf{D})=\mathbf{0}_{3\times1} \quad (44)$$

where $\mathbf{A}_{init}\triangleq2\left[\mathbf{F}^T(\tau_0)\otimes\left[\mathbf{q}_{n,0}^{b}{}^{*}\right]^{-}\right]_{2:4}$. Assume the initial biases are roughly known, the coefficient $\mathbf{D}$ can be obtained by using all available measurements and the initial attitude in the current time window as

$$\min_{\mathbf{D}}\left\|\mathbf{A}\,vec(\mathbf{D})\right\| \quad \text{s.t. } \left\|\mathbf{B}\,vec(\mathbf{D})\right\|=1 \quad (45)$$

where the constraint term denotes the unit norm of the initial quaternion, which helps ensure a non-zero solution. The matrices $\mathbf{A}$ and $\mathbf{B}$ are known matrices as follows

$$\mathbf{A}=\left[\mathbf{A}_{init}^T \quad \mathbf{A}_g^T \quad \mathbf{A}_a^T \quad \mathbf{A}_m^T\right]^T \quad (46)$$
$$\mathbf{B}=\mathbf{F}^T(\tau_0)\otimes\mathbf{I}_4$$

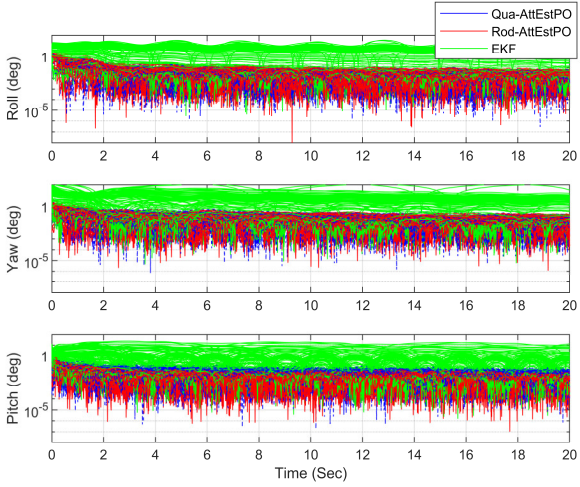

Fig. 2. Euler angle errors of Qua-AttEstPO, Rod-AttEstPO and EKF across 100 Monte-Caro runs.

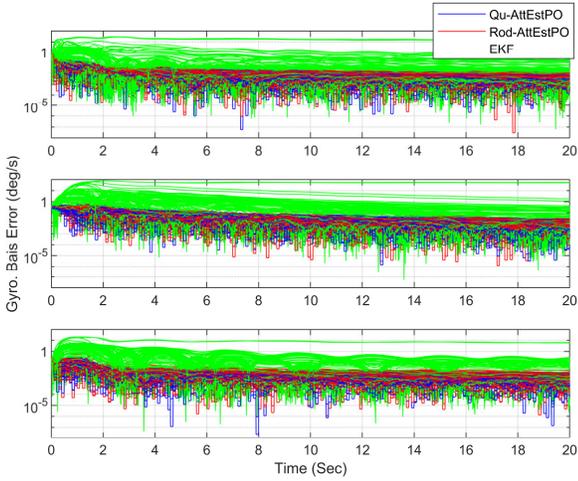

Fig. 3. Gyroscope bias errors of Qua-AttEstPO, Rod-AttEstPO and EKF across 100 Monte-Caro runs.

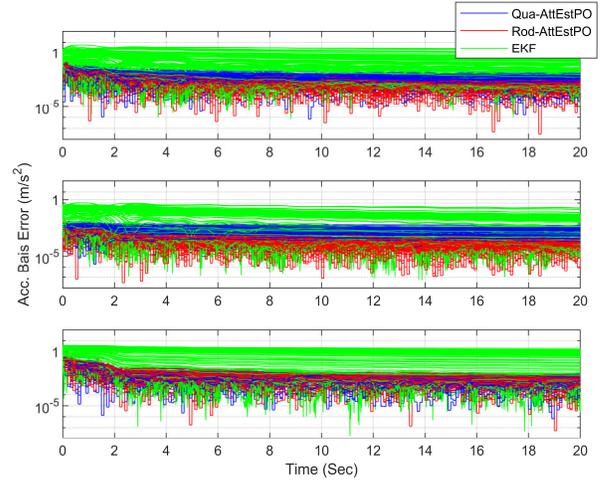

Fig. 4. Accelerometer bias errors of Qua-AttEstPO, Rod-AttEstPO and EKF across 100 Monte-Caro runs.

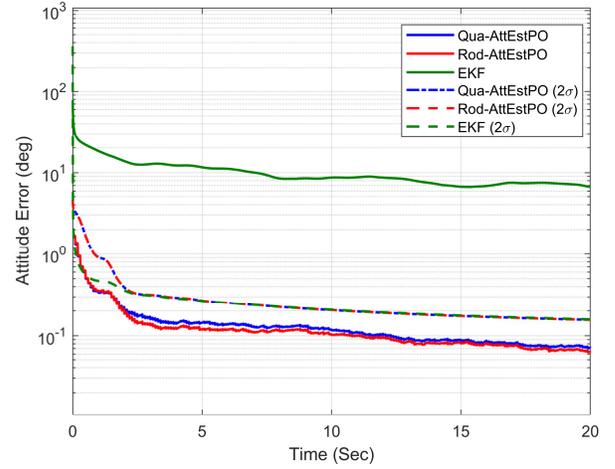

Fig. 5. Average attitude errors across 100 Monte-Caro runs for Qua-AttEstPO, Rod-AttEstPO and EKF with the associated 2-standard deviation bound.

where $\mathbf{A}_g \triangleq \left[ \mathbf{a}_g^T(\tau_0) \; \ldots \; \mathbf{a}_g^T(\tau_N) \right]^T$ is the matrix formed by the fitted angular velocity at Chebyshev points, $\mathbf{A}_a \triangleq \left[ \mathbf{a}_a^T(t_j) \; \mathbf{a}_a^T(t_k) \; \cdots \right]^T$ and $\mathbf{A}_m \triangleq \left[ \mathbf{a}_m^T(t_l) \; \mathbf{a}_m^T(t_p) \; \cdots \right]^T$ respectively denote the matrices from accelerometer and magnetometer measurements and the timestamp in $\mathbf{a}_m$ and $\mathbf{a}_a$ denotes all the feasible measurement times that are detected by (9) and (10). The minimization in (45) is a homogenous least squares problem of the quaternion Chebyshev coefficients, which can be solved by the generalized singular value decomposition [42].

As for the attitude estimation problem in the form of the Rodrigues vector in (32), the initial coefficients can be readily obtained by transforming the obtained quaternion coefficients in (45) to the corresponding Rodrigues vector update coefficients as

$$\mathbf{h}_i \approx \frac{2 - \delta_{0i}}{P_{N_g}} \sum_{k=0}^{P_{N_g}-1} \Delta \mathbf{g} \left( \Delta \mathbf{q} \left( \cos \frac{(k+1/2)\pi}{P_{N_g}} \right) \right) \cos \frac{i(k+1/2)\pi}{P_{N_g}}$$

(47)

where $\delta_{0i}$ is the Kronecker delta function, yielding 1 for $i = 1$ and zero otherwise. The exact coefficients could be obtained only if the number of summation terms, $P_{N_g}$, approaches infinity [35]. $\Delta \mathbf{g}(\Delta \mathbf{q}(\tau))$ transforms the quaternion update to the Rodrigues vector update as shown in (15), among which $\Delta \mathbf{q}(\tau)$ is obtained from (13) using the resultant $\mathbf{q}_a^b(\tau)$ by (45).

Hereafter, we name the proposed attitude quaternion and Rodrigues vector estimation by polynomial optimization as Qua-AttEstPO and Rod-AttEstPO, respectively. Table I lists their main steps. The algorithm input includes the user-defined time-window size, the Chebyshev polynomial order for attitude approximation, the feasible sensor measurements in the current window and the initial state with the associated covariance. After the estimation is completed, the state and covariance at the end of current time window are exploited as the initial state and covariance for the next time window.

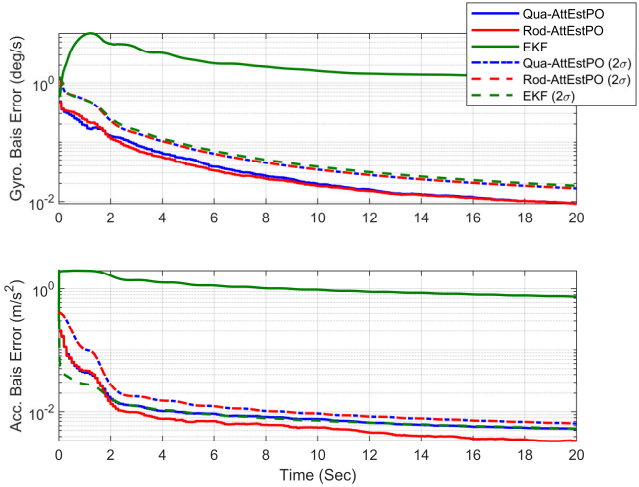

Fig. 6. Average gyroscope and accelerometer bias errors across 100 Monte-Caro runs for Qua-AttEstPO, Rod-AttEstPO and EKF with the associated 2-standard deviation bounds.

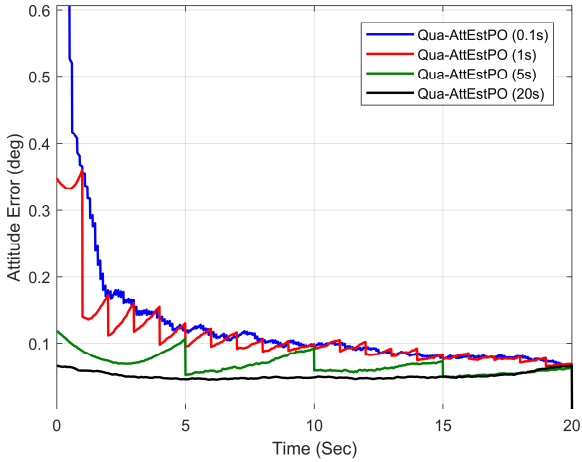

Fig. 7. Average attitude errors across 100 Monte-Caro runs for Qua-AttEstPO with different window sizes.

## IV. SIMULATIONS AND EXPERIMENTS

### A. Simulation Results

In this section, simulations are conducted to evaluate the proposed Qua-AttEstPO and Rod-AttEstPO algorithms, against the popular multiplicative EKF. The estimate state in EKF includes the attitude, gyroscope and accelerometer biases. Interested readers are referred to [7] for more details about the EKF implementation.

Assume a 9-axis MIMU rotating at a fix position (longitude: 112 deg, latitude: 28 deg height: 0 m) under the classic coning motion for 20 seconds. The attitude time trajectory is described by the quaternion as $\mathbf{q}_n^b = \cos(\alpha/2) + \sin(\alpha/2)\begin{bmatrix} 0 & \cos(\varsigma t) & \sin(\varsigma t) \end{bmatrix}^T$, where the coning frequency $\varsigma = 0.74\pi$ rad/s and the coning angle $\alpha = 10$ deg. The sampling frequency of the MIMU is set to 100Hz. The biases of gyroscopes and accelerometers are assumed to be $\begin{bmatrix} 0.5 & 0.3 & -0.2 \end{bmatrix}^T$ deg and $\begin{bmatrix} 0.1 & 0.2 & -0.2 \end{bmatrix}^T$ m/s$^2$, respectively. The root power spectral density of the gyroscope measurement noise of each axis is set to $1°/\sqrt{h}$, and the noise standard deviations of the normalized magnetometer and accelerometer measurements in each axis are set to 0.02 and 0.01 m/s$^2$, respectively.

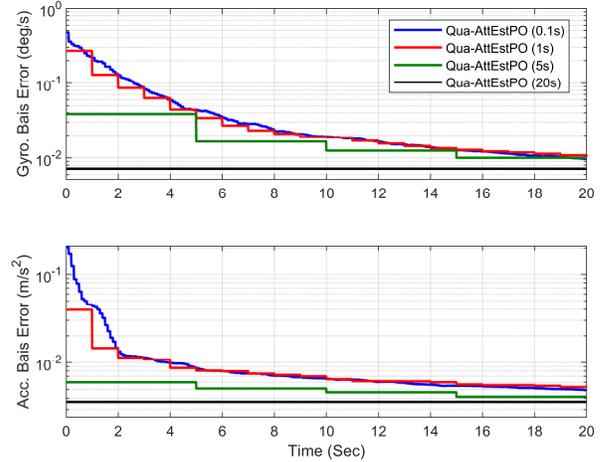

Fig. 8. Average gyroscope and accelerometer bias errors across 100 Monte-Caro runs for Qua-AttEstPO with different window sizes.

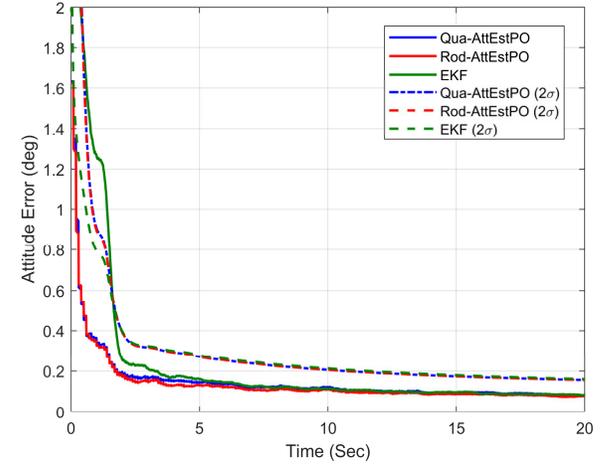

Fig. 9. Average attitude errors across 100 Monte-Caro runs for Qua-AttEstPO, Rod-AttEstPO and EKF with the associated 2-standard deviation bounds for small initial attitude error.

TABLE II
RMSE OF FINAL ATTITUDE ESTIMATES

|  | Roll (deg) | Yaw (deg) | Pitch (deg) |
|---|---|---|---|
| EKF | 4.884 | 12.364 | 2.233 |
| Qua-AttEstPO | 0.026 | 0.082 | 0.026 |
| Rod-AttEstPO | 0.026 | 0.077 | 0.015 |

For all algorithms, the initial gyroscope and accelerometer biases are set to zeros and the statistics of the random noises are assumed to be known. The Chebyshev polynomial orders of both quaternion and Rodrigues vector approximation are set to $N_q = N_g = 6$, the time window size is set to 0.1 seconds and the number of Chebyshev points in the window is set to $N + 1 = 7$. In practical scenarios, the initial level angle (roll and pitch) can be readily obtained by the accelerometer leveling, but the yaw is more difficult to determine by the magnetometer due to the magnetic disturbance nearby. Thus, a 180 deg initial yaw error is assumed to fully assess the performance of different algorithms. Specifically, the initial orientation errors of roll, yaw and pitch are assumed as Gaussian noises with zero mean and the standard derivation of $[5\ 180\ 5]$ deg .

| Algorithm (Window Size) | Rod-AttEstPO (0.1s) | Qua-AttEstPO (0.1s) | Qua-AttEstPO (1s) | Qua-AttEstPO (5s) | Qua-AttEstPO (20s) | EKF |
|---|---|---|---|---|---|---|
| Time Cost | 0.91 s | 3.37 s | 4.93 s | 25.4 s | 52. 5 s | 0.35 s |

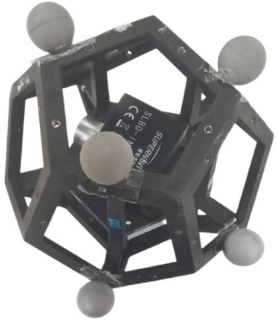

Fig. 10. The MIMU installation in the experiment. The MIMU is placed in the center of the dodecahedron and five Vicon markers are attached on the edges to acquire the true attitude of the dodecahedron.

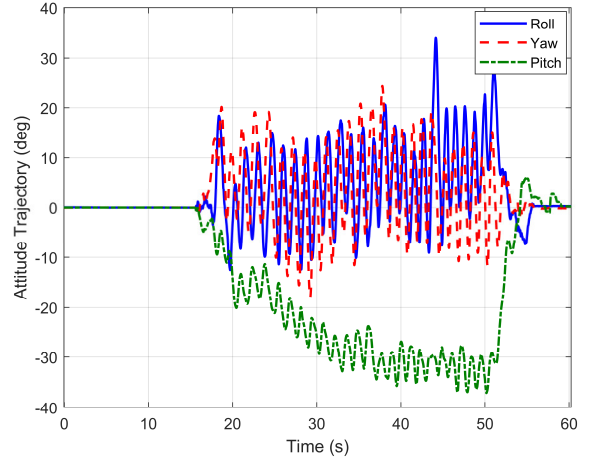

Fig. 11. The rough attitude trajectory obtained from the integration of the calibrated gyroscope measurements.

Figures 2-4 plot the attitude, gyroscope and accelerometer bias errors across 100 Monte Carlo runs. We see that the Qua-AttEstPO and Rod-AttEstPO are much superior to EKF in terms of stability and convergence speed. The average of absolute estimation errors is defined to quantity the estimation accuracy as

$$\varepsilon_i(k) = \frac{1}{L}\sum_{l=1}^{L}\left\|\hat{\mathbf{x}}_{(i),k}^{(l)} - \mathbf{x}_{(i),k}^{(l)}\right\| \tag{48}$$

where $L$ denotes the number of Monte Carlo runs, and $i$ and $k$ denote the $i$-th state component at time $k$. For brevity, Figs. 5-6 respectively plot the norm of the average errors for attitude and gyroscope/accelerometer biases. The $2\sigma$ bounds are calculated by twice the averaged square root of diagonal elements of the covariance matrix. For a consistent filter, the estimation error should stay below the $2\sigma$ derivation bounds with a possibility of 95% [43]. Figures 5-6 indicate that both Qua-AttEstPO and Rod-AttEstPO are consistent estimators, while the EKF is too optimistic. The RMSE of the estimate at the end of simulation is listed in Table II. It shows that the Rod-AttEstPO is marginally better than Qua-AttEstPO, which may be arguably owed to the fact that the parameter dimension of Qua-AttEstPO is higher than that of Rod-AttEstPO and solving the constrained optimization in Qua-AttEstPO may be less accurate than the unconstrained optimization in Rod-AttEstPO. However, due to the singularity of the Rodrigues vector at a $180°$ rotation, it is not recommended to utilize the Rod-AttEstPO with a large window size. Therefore, we only use Qua-AttEstPO for a large window computation in the sequel. Figures 7-8 plot the average attitude and sensor bias errors of Qua-AttEstPO with different window sizes, namely, 0.1/1/5/20 seconds. The Chebyshev polynomial orders are respectively set to 6, 40, 150, and 300. We see that the estimation accuracy improves along with the increased window size. Larger window size means the algorithm employs more sensor information to estimate the state, which causes a larger time delay. Since the algorithm cannot execute until all the measurements in the current window come in. Note that the generalized singular value decomposition of a high-dimension matrix as in (45) is time-consuming. Therefore, the initialization of Chebyshev coefficients for AttEstPO in a larger window is carried out by fitting the states acquired by AttEstPO of 0.1s window size.

All of the proposed algorithms are implemented with the function 'lsqnonlin' on the MATLAB platform. The average time cost across 100 Monte Carlo runs is listed in Table III. The time costs of Rod-AttEstPO and Qua-AttEstPO with 0.1s window size are respectively about 3 and 10 times larger than that of EKF, which can ensure a real-time algorithm execution for 20-second data. Although the computation burden of Qua-AttEstPO increases dramatically along with the window size, it can still be used for batch estimation.

The average attitude error across 100 Monte Caro simulation runs for small initial attitude error is plotted in Fig. 9. The initial attitude error is presumed as Gaussian noise with zero mean and $[5\ 10\ 5]$ deg standard deviation, the window size is 0.1s and the Chebyshev polynomial order is 6, and other settings are the same with the above. The three algorithms achieve similar final accuracy, but the convergence speed of Qua-AttEstPO and Rod-AttEstPO is faster than that of EKF and the EKF appears inconsistent either in the first several seconds.

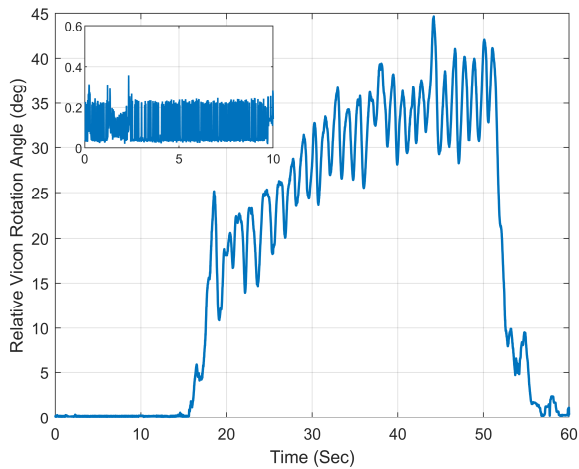

Fig. 12. Relative rotation angle of the reference Vicon system in the experiment (Upper left is the details of the first 10 seconds).

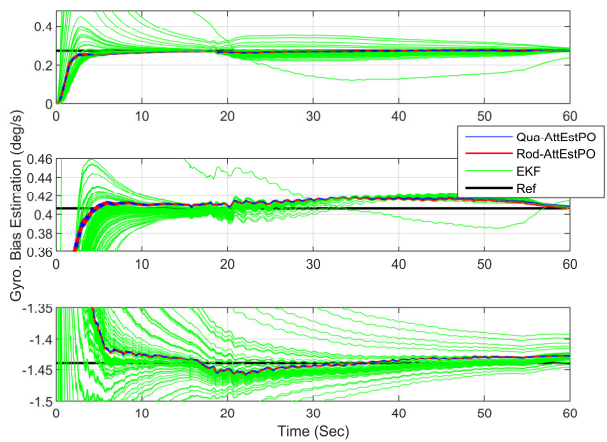

Fig. 14. Gyroscope bias estimate results across 100 Monte-Caro runs for Qua-AttEstPO, Rod-AttEstPO, EKF and the reference in the experiment.

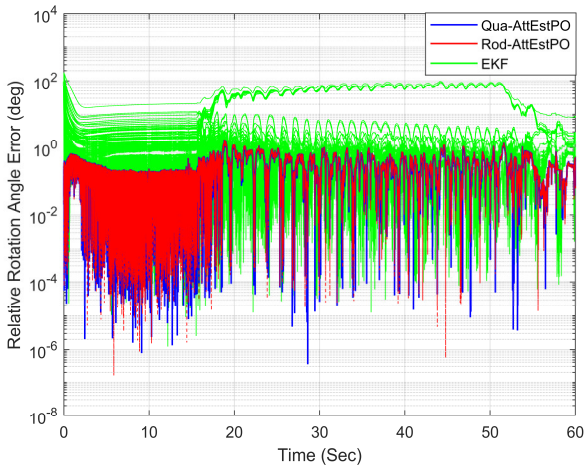

Fig.13. Relative rotation angle errors across 100 Monte-Caro runs for Qua-AttEstPO, Rod-AttEstPO and EKF in the experiment.

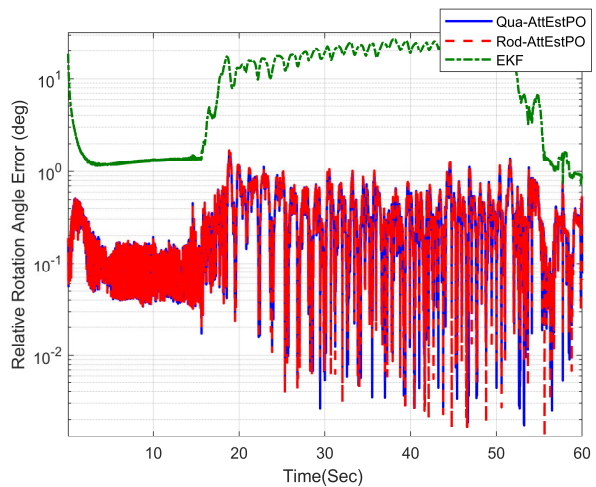

Fig. 15. Average of relative rotation angle errors across 100 Monte Carlo runs for Qua-AttEstPO, Rod-AttEstPO and EKF in the experiment.

### B. Experiment Results

The experiment is conducted using the Xsens MTW unit with the sample frequency of 100 Hz. The bias stabilities of gyroscope and accelerometer are 10 deg/h and 0.1 mg, respectively. The power spectrum densities of gyroscope, accelerometer and magnetometer noises are respectively $0.01 \text{ deg/s/}\sqrt{\text{Hz}}$, $20 \text{ μg/}\sqrt{\text{Hz}}$ and $0.2 \text{ mGauss/}\sqrt{\text{Hz}}$. A Vicon motion capture system is used to provide the true attitude relative to its inherently-defined reference frame. As shown in Fig. 10, the MIMU is fixed at the center of a regular dodecahedron, with five Vicon markers attached on the edges of the dodecahedron. The Vicon system tracks the attitude change by the markers. In view of the magnetic material in the dodecahedron, the magnetometer is calibrated before the experiment [32, 33]. In the experiment, the unit stays stationary on a bench for about 90 seconds, then is picked up and rotated around itself for about 40 seconds, and finally put back on the bench at the same attitude. Two datasets are collected to test the proposed algorithms. The two datasets produce similar results, so we only show one dataset result in this paper for brevity. The rough attitude time trajectory is plotted in Fig. 11, which is obtained by integrating the gyroscope measurements with the gyroscope bias roughly calibrated by averaging the static gyroscope outputs before the rotation.

Due to the attitude misalignment between the MIMU and dodecahedron, their attitudes are not exactly the same. However, the relative rotation angle of the MIMU body frame, denoted by $\alpha_{b(0)}^{b(t)}$, should be identical with that of the Vicon. Specifically, the attitude quaternion estimate $\mathbf{q}_{n(t)}^{b(t)}$ of the MIMU is transformed to the relative attitude quaternion $\mathbf{q}_{b(0)}^{b(t)}$ by multiplying the conjugate of the MIMU initial quaternion $\mathbf{q}_{n,0}^{b}$, i.e., $\mathbf{q}_{b(0)}^{b(t)} = \mathbf{q}_{n,0}^{b}{}^{*} \circ \mathbf{q}_{n}^{b}(t)$. Then, the relative rotation angle can be computed by

$$\alpha_{b(0)}^{b(t)} = 2\arccos\left(\left[\mathbf{q}_{b(0)}^{b(t)}\right]_1\right) \tag{49}$$

where $\left[\mathbf{q}_{b(0)}^{b(t)}\right]_1$ denotes the first element of $\mathbf{q}_{b(0)}^{b(t)}$ and the relative angle is defined on $\begin{bmatrix} 0 & \pi \end{bmatrix}$. The error of the relative rotation angle is defined as

$$\Delta\alpha = \left\| \alpha_{vicon} - \alpha_{b(t)}^{b(t)} \right\| \tag{50}$$

where $\alpha_{vicon}$ is the relative rotation angle of the Vicon system, which is similarly computed from the Vicon's attitude output. Figure 12 plots the relative rotation angle from the Vicon system, where the variation during the first 10-second static outputs indicates that the accuracy of relative rotation angle is about 0.3 deg.

The window size of Rod-AttEstPO and Qua-AttEstPO is 0.1s and the Chebyshev polynomial order is 6. A set of Monte Carlo runs are also performed in the experiment by intentionally randomizing the initial conditions. Specifically, the initial gyroscope and accelerometer biases are set to zeros and the initial attitude error is set to Gaussian noise with zero mean and $\begin{bmatrix} 5 & 180 & 5 \end{bmatrix}$ deg standard deviation. Figures 13-14 plot the relative rotation angle error and the gyroscope bias across 100 Monte Carlo runs, where the reference gyroscope bias is roughly obtained by averaging the static gyroscope outputs before the rotation. It can be seen that the results of Qua-AttEstPO and Rod-AttEstPO for different initial state values are quite similar in each run, which demonstrates the superior stability of AttEstPO in face of large initial attitude errors. And, the average of relative rotation angle errors in Fig. 15 also highlights the accuracy superiority of Qua-AttEstPO and Rod-AttEstPO over EKF. The RMSEs of relative rotation angle at the end of experiment are about 1.1, 0.03, 0.02 degrees for EKF, Qua-AttEstPO and Rod-AttEstPO, respectively.

## V. CONCLUSION

Attitude estimation by way of inertial-magnetic sensors is vital for many applications. Traditional linearization-based EKF is prone to divergence in the challenging scenarios with large initial state error. To improve the estimation stability and convergence speed, this paper introduces the Chebyshev collocation method into the attitude estimation, resulting in the so-called Qua-AttEstPO and Rod-AttEstPO algorithms. They essentially transform the quaternion/Rodrigues vector estimation into constrained/unconstrained least squares problem in a time window of interest. Due to the inherent singularity of the Rodrigues vector, Rod-AttEstPO is only suitable for a short time window. Simulation and field test results show that the Qua-AttEstPO and Rod-AttEstPO algorithms have similar accuracy with each other and perform quite well under large initial errors. Increasing the size of time window in Qua-AttEstPO can further improve the estimation accuracy yet at the price of time delay. Therefore, Qua-AttEstPO with a large window size can be treated as a batch attitude estimation method. Our future work will explore the advantage of the polynomial optimization-based estimation in other navigation applications, e.g., INS/GNSS integrated navigation.

## ACKNOWLEDGMENT

Thanks to Miss Miaomiao Yan and Dr. Boying Li, at Shanghai Jiao Tong University, for their test assistance with the Vicon tracking system.